\documentclass[10pt,twocolumn,letterpaper]{article}

\usepackage{cvpr}              % To produce the CAMERA-READY version
\usepackage{graphicx}
\usepackage{amsmath}
\usepackage{amssymb}
\usepackage{booktabs}
\usepackage{xcolor}
\usepackage{verbatim}
\usepackage{rotating}
\usepackage[symbol]{footmisc}

\usepackage[pagebackref,breaklinks,colorlinks]{hyperref}
\DeclareMathOperator*{\argmin}{arg\,min}

\usepackage[capitalize]{cleveref}
\crefname{section}{Sec.}{Secs.}
\Crefname{section}{Section}{Sections}
\Crefname{table}{Table}{Tables}
\crefname{table}{Tab.}{Tabs.}

\def\cvprPaperID{*****} % *** Enter the CVPR Paper ID here
\def\confName{CVPR}
\def\confYear{2022}

\begin{document}

%%%%%%%%% TITLE - PLEASE UPDATE
\title{GaLeNet: Multimodal Learning for Disaster Prediction, Management and Relief}

\begin{comment}
\author{Rohit Saha\qquad
        Mengyi Fang\qquad
        Angeline Yasodhara\qquad
        Kyryl Truskovskyi\qquad
        Azin Asgarian\textsuperscript{\textdagger}
        \\\textbf{Georgian}\\
        \{rohit.saha, mengyi.fang, angeline, kyryl, azin\}@georgian.io\\ \\
        Daniel Homola\qquad
        Raahil Shah\qquad
        Frederik Dieleman\qquad
        Jack Weatheritt\qquad
        Thomas Rogers\textsuperscript{\textdagger}
        \\\textbf{Tractable}\\
        \{daniel.homola, raahil.shah, frederik.dieleman, jack.weatheritt, thomas.rogers\}@tractable.ai\\
        }
\end{comment}

\author{Rohit Saha\textsuperscript{1}
    \and \hspace{-0.1in}Mengyi Fang\textsuperscript{1}
    \and \hspace{-0.1in}Angeline Yasodhara\textsuperscript{1}
    \and \hspace{-0.1in}Kyryl Truskovskyi\textsuperscript{1}
    \and \hspace{-0.1in}Azin Asgarian\textsuperscript{1,\textdagger}
    \and \vspace{0.05in}
    \and \hspace{-0.3in}Daniel Homola\textsuperscript{2}
    \and \hspace{-0.1in}Raahil Shah\textsuperscript{2}
    \and \hspace{-0.1in}Frederik Dieleman\textsuperscript{2}
    \and \hspace{-0.1in}Jack Weatheritt\textsuperscript{2}
    \and \hspace{-0.1in}Thomas Rogers\textsuperscript{2,\textdagger}
}

\maketitle
\def\thefootnote{*}\footnotetext{All authors have contributed equally to this work.}
\def\thefootnote{\textdagger}\footnotetext{Correspondences: azin@georgian.io, thomas.rogers@tractable.ai}

%============================================================
% Abstract
%============================================================
\begin{abstract}
After a natural disaster, such as a hurricane, millions are left in need of emergency assistance. To allocate resources optimally, human planners need to accurately analyze data that can flow in large volumes from several sources. This motivates the development of multimodal machine learning frameworks that can integrate multiple data sources and leverage them efficiently. To date, the research community has mainly focused on unimodal reasoning to provide granular assessments of the damage. Moreover, previous studies mostly rely on post-disaster images, which may take several days to become available. In this work, we propose a multimodal framework (GaLeNet) for assessing the severity of damage by complementing pre-disaster images with weather data and the trajectory of the hurricane. Through extensive experiments on data from two hurricanes, we demonstrate (i) the merits of multimodal approaches compared to unimodal methods, and (ii) the effectiveness of GaLeNet at fusing various modalities. Furthermore, we show that GaLeNet can leverage pre-disaster images in the absence of post-disaster images, preventing substantial delays in decision making.
\end{abstract}

%============================================================
% Introduction
%============================================================
\section{Introduction}

In Haiti alone, Hurricane Matthew left an estimated 180,000 people homeless and 1.4 million in need of emergency assistance~\cite{rodriguez2016aftermath}. When so many people are affected, it is critical to get an accurate assessment of the location and severity of damage so that resources can be allocated quickly to where they are needed most. The earlier this picture of destruction can be  assembled, the better a natural disaster can be managed. Disaster management and prediction could become even more important in the future, as the frequency of natural disasters might be increasing due to climate change~\cite{banholzer2014impact}.

One way to identify affected areas is to use post-disaster (``gray skies'') images captured by satellite, drone, or plane in the immediate aftermath of a disaster. However, there can be a delay of three days before such images are available~\cite{Voigt2007}. There is a multitude of data that can be gathered before or during the event such as pre-disaster (``blue skies'') images, weather features or dynamics, and the trajectory of the hurricane. Are we able to reason over such a diverse range of data modalities to predict which buildings will be affected most before we have direct visual evidence of the destruction?

Recently, several works have been published on applying machine learning to natural disaster prediction and management by predicting the location and severity of damage to buildings~\cite{xu2019building,shen2020cross,oludare2021semi,gupta2021rescuenet,wu2021building,bai2020pyramid,weber2020building}. However, these rely on direct visual evidence of the damage provided by the post-disaster images. There are also an increasing number of publications on multimodal machine learning, but these use modalities that are naturally aligned (e.g. twitter images and text) and typically only provide ``big picture''~\cite{algiriyage2022multi} insights about the damage such as classifying image-text pairs from social media for their informativeness, type of emergency and severity of emergency~\cite{abavisani2020multimodal}. 

In this work, we propose a multimodal framework to predict the severity of the damage \emph{before} images of the damage are available. The motivation is to allow for early triage and allocation of resources. We also show how the accuracy of the framework improves once the post-disaster images are collected. This could allow authorities to adapt their relief program as new information becomes available.

\def\thefootnote{1}\footnotetext{\textbf{Georgian}. Toronto, Canada.}
\def\thefootnote{2}\footnotetext{\textbf{Tractable}. London, United Kingdom.}

\vspace{-0.1in}
\renewcommand{\thefootnote}{\fnsymbol{footnote}}
\paragraph{Contributions} To our knowledge, this work is the first to attempt using multimodal machine learning to build a granular picture of damage severity without using post-disaster imaging. This is achieved by: (i) aligning the data modalities via the longitude, latitude and time of event, and (ii) supplementing pre-disaster images with weather data as well as information about the hurricane trajectory through a novel featurization method. The framework also allows for the assessment to be updated once post-disaster imagery is available. Through the use of pre-trained image embeddings, we are able to efficiently train our framework on a limited set of examples. We call our framework GaLeNet\footnote[2]{Coined by conjoining the words ``Gale'' and ``LeNet''.}. 

%============================================================
% Related Work
%============================================================
\section{Related work}
The need for rapid damage assessment following a natural disaster has driven research into machine learning methods that can quickly process data to assess the location and severity of damage~\cite{algiriyage2022multi}. Decision makers are often faced with the challenge of promptly integrating several streams of information, reasoning over them, and coming to an accurate assessment~\cite{chamola2020disaster} which may evolve as more data becomes available.

Multimodal learning is a natural solution to this problem; however, work in this area is still limited to understanding damage at a high level~\cite{algiriyage2022multi}. Methods capable of making granular inferences have so far been limited to unimodal image-based methods that are mostly single task ~\cite{xu2019building,shen2020cross,oludare2021semi,bai2020pyramid,wu2021building,weber2020building} and sometimes multitask~\cite{gupta2021rescuenet} - all of which rely on the post-disaster image.

Despite the recent advances in deep learning, the applications of these models for natural disaster prediction is still limited due to the scarcity of training data. This is even more severe in a multimodal setting where several data sources have to be sourced and merged. This has limited the application of deep learning, although these methods are generally effective given enough data~\cite{arinta2019natural}. Here, we overcome this problem by using pre-trained models as feature extractors (for images) and handcrafted representations of the hurricane trajectory and weather data leading up to the event. For images, we observe that using multiple scales helps with the representation, an observation which motivated~\cite{bai2020pyramid}. 

For the hurricane trajectory, we present a novel featurization method based on calculating the closest point along the hurricane trajectory. The authors are also not aware of any building damage prediction models that use the hurricane trajectory as an input modality.

Aligning separate streams of data is one of the challenges of multimodal learning~\cite{baltruvsaitis2018multimodal}. In this work, we align data using the geolocation (longitude and latitude) and timestamp of the disaster.

%============================================================
% Methods
%============================================================
\section{Methods}
\subsection{Problem Statement \& Scope}
We consider both a \emph{proactive} and a \emph{reactive} scenario. In the \emph{reactive} case, the goal is to predict which buildings will be damaged, and the severity of the damage, using direct visual evidence collected through post-disaster satellite imaging. However, in the \emph{proactive} case, we are only allowed to use contextual information such as pre-disaster imaging or weather data.

In this work, we limit the scope to include buildings in the vicinity of hurricanes Matthew (2016)~\cite{hurricaneMatthew} and Michael (2018)~\cite{hurricaneMichael}. We focus on hurricanes as there is more contextual information (e.g. hurricane trajectory or weather data) than is available for other disasters such as forest fires, thus making the problem more compatible with multimodal learning.

We treat the problem as a classification task, where for each building, we classify the damage into one of four levels of severity: (i) no damage, (ii) minor damage, (iii) major damage, and (iv) destroyed. The labels and class definitions are taken from the xBD dataset~\cite{gupta2019xbd}. Other than selecting only examples related to wind damage due to hurricanes, we used the original train, test, and hold splits from the xView 2 challenge, throughout.

\subsection{Data Modalities}
The xBD dataset~\cite{gupta2019xbd} was used as the foundation of our multimodal dataset, as it contains labels for the severity of damage as well as the geolocation and time of the disasters. We used these as anchor points to align all of our different modalities: (i) pre-disaster satellite images, (ii) post-disaster satellite images, (iii) weather data, and (iv) the hurricane trajectory.

After filtering the original xBD dataset~\cite{gupta2019xbd} for hurricanes and wind damage, 36,625, 9,283, and 12,791 building were identified for training, validation and testing. The dataset is somewhat imbalanced with 48\% of the labels being no damage and the remaining 52\% consisting of the other three levels of damage (33\% minor damage, 11\% major damage, and 8\% destroyed). 

\vspace{-.2cm}
\subsubsection{Image Data}
\vspace{-.1cm}
\paragraph{Source} The images from the xBD dataset~\cite{gupta2019xbd} were pre-processed by taking the centroid of each building polygon and then taking a centered crop, as opposed to using the full satellite image. 

\vspace{-.4cm}
\paragraph{Representation} We tried both uniscale and multiscale approaches for visual representation. A zoomed out view (denoted by Scale-1x) was created by scaling the image by $s{\approx}11/\sqrt{A}$ (where $A$ is the area of the building polygon) in each dimension before performing a crop, centered around the building center, of $224{\times}224$ pixels. Additional crops were performed by increasing the scale to $4s$ (Scale-4x), $16s$ (Scale-16x), and $32s$ (Scale-32x) - progressively zooming in on the building.

Visual features were extracted using: (i) a pre-trained CNN baseline, (ii) the CLIP~\cite{radford2021learning} visual branch, and (iii) a pre-trained U-Net on satellite images to segment buildings by type (houses, building, and sheds/garages)~\cite{obravo7}. The CNN baseline consists of five convolutional blocks followed by two fully-connected layers. We trained this model on the cropped images from the xBD training set for the damage classification task. We explored four different versions of CLIP~\cite{radford2021learning} for extracting visual features: (i) CLIP ViT-B/32 (ii) CLIP ViT-B/16 (iii) CLIP ViT-L/14 and (iv) CLIP ViT-L/14@336px. For the U-Net, we used the pre-trained weights available from~\cite{obravo7}.

\vspace{-.2cm}
\subsubsection{Hurricane Trajectory}
\vspace{-.1cm}
\paragraph{Source} The trajectory of the two hurricanes were extracted from reports published by the NOAA~\cite{hurricaneMatthew, hurricaneMichael}. The trajectory is a series of longitude and latitude coordinates with an associated time, wind speed and central pressure.

\vspace{-.4cm}
\paragraph{Representation} To featurize the hurricane trajectory, we used the haversine formula to compute the shortest distance between each building and the trajectory. This shortest distance (in km) and the wind speed and pressure at this point in the trajectory were extracted. In case of ties (i.e. multiple points on the trajectory that had the same distance to the building), the maximum wind speed and pressure were used. We also experimented with more complex multiscale and multipoint schemes as well as distance-weighted features, but these offered no benefit.
% We denote this representation as NOAA v1. In the second approach, we took the minimum, average, and standard deviation of the wind speed and pressure at the nearest point on the trajectory. In addition, we also computed the same metrics as well as maximum and distance-weighted average of the wind speed and pressure at trajectory spots within 5, 20, 50, and 100 km of the building. Lastly, we also calculated the distance-weighted average of the wind speed and pressure along all trajectory spots. We call this representation NOAA v2.

\vspace{-.2cm}
\subsubsection{Weather Data}
\vspace{-.1cm}
\renewcommand{\thefootnote}{\fnsymbol{footnote}}
\paragraph{Source} 
The weather data were extracted from OpenWeather\footnote[2]{Data was collected using the OpenWeather API \url{https://openweathermap.org/}} and consists of many factors including, but not limited to, the temperature, wind speed and direction, precipitation, humidity, pressure, and visibility. We collected daily weather data for seven days preceding each hurricane, with eight time points per day.

\vspace{-.3cm}
\paragraph{Representation} The best representation of the weather data was simply the average of each weather feature across all time points. We also tried using off-the-shelf feature engineering libraries for time series, such as TSFresh~\cite{christ2018time} and Catch22~\cite{lubba2019catch22}, but these offered limited benefit and resulted in a substantial increase in the number of input features and model parameters, making it prone to overfitting.

\begin{figure}[tbh] 
 \centering
 \includegraphics[width=.4\paperwidth]{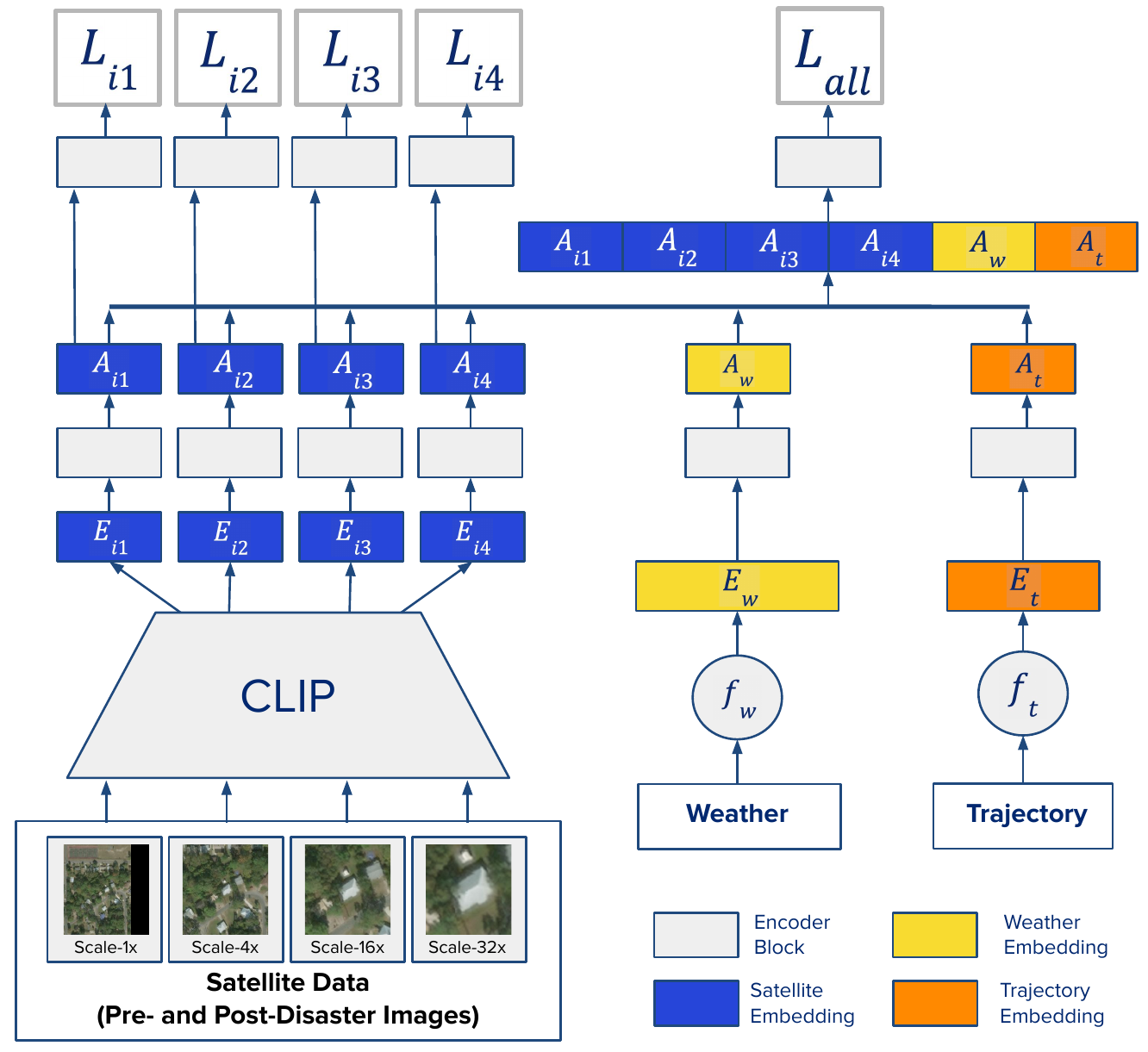}
\caption{Schematic diagram of our proposed framework}
\vspace{-.3cm}
\label{fig:ram-net_architecture}
\end{figure}

\subsection{Metrics}
The following metrics were used for evaluation: (i) Area Under the Precision Recall Curve (PR AUC), (ii) Area Under the ROC Curve (ROC AUC), and (iii) Balanced Accuracy (Bal. Acc.) which is the unweighted average of recall obtained on each class. Since we are dealing with an imbalanced dataset, we use macro averaging to obtain the PR AUC and ROC AUC curves.

\section{Experiments}
\label{sec:experiments}
To study the contribution of each modality, we conduct two sets of experiments. First, we evaluate and compare the representations of each modality separately on damage classification. Then, using the best representation of each modality, we assess the effectiveness of our multimodal framework for both the proactive and reactive scenarios.

\subsection{Single Modalities}
\label{subsec:single_modalities}
To evaluate and compare the different modalities and their representations, we follow the common practice~\cite{kolesnikov2019revisiting} and train a linear logistic regression (LogReg) model for damage severity classification. We train LogReg using the L-BFGS~\cite{liu1989limited} solver. We choose the best value for the inverse regularization strength $C$ by running a grid search over $C{\in}10^{[-3, 3]}$ on the validation data.

\subsection{Multiple Modalities}
\label{subsec:multiple-modalities}
We assess the effectiveness of our multimodal framework for proactive and reactive scenarios by comparing it against two models: (i) LogReg, and (ii) Concat-MLP. For LogReg, we use the setup described in \ref{subsec:single_modalities}. In the reactive case, we use the hurricane trajectory, weather data, and the post-disaster satellite images. In the proactive case, however, we use pre-disaster images instead of the post-disaster images.

\vspace{-.2cm}
\paragraph{Concat-MLP}
We begin by concatenating the best representations of each modality, namely, the embeddings extracted from CLIP ViT-L/14@336px corresponding to the four different image scales, denoted by $\{E_{i1}, E_{i2}, E_{i3}, E_{i4}\}$, the weather embedding $E_w$, and the trajectory embedding $E_t$. Let $E_{all}{=}[E_{i1}, E_{i2}, E_{i3}, E_{i4}, E_w, E_t]$ be the concatenated representation. Next, $E_{all}$ is passed through a network of two fully-connected layers. The first layer contains 128 nodes, followed by 32 nodes in the second layer, and both layers use ReLU ~\cite{agarap2018relu} activation. The output of the network is fed into a softmax layer for classification.

\vspace{-.2cm}
\paragraph{GaLeNet} 
We observe that an early-stage naïve concatenation of feature embeddings, extracted from different modalities, can lead to difficulties in training due to: (i) different modalities having variable rates of learning~\cite{wang2019multimodalhard}, and (ii) modalities containing different levels of information for the task at hand. To combat these problems, we use \textit{late-fusion}; a common paradigm in which modality-specific latent representations are learned to aid the process of fusion \cite{owens2018audiovisual, Zhang2019ALF, ju2017ensemble}.

To this end, our proposed framework (Figure~\ref{fig:ram-net_architecture}) jointly trains multiple modality-specific encoders, whose
intermediate activations are then concatenated. Each encoder follows a similar sequence of computations; \textit{linear projection} $\rightarrow$ \textit{batch normalization} $\rightarrow$ \textit{ReLU} $\rightarrow$ \textit{dropout}. The CLIP embeddings $\{E_{i1}, E_{i2}, E_{i3}, E_{i4}\}$ are passed through their corresponding encoders to yield $\{A_{i1}, A_{i2}, A_{i3}, A_{i4}\} \in \mathbb{R}^{56}$. Similarly, the encoders corresponding to $E_w$ and $E_t$ produce $A_w \in \mathbb{R}^{16}$ and, $A_t \in \mathbb{R}^{3}$ respectively. Next, these intermediate activations are concatenated, yielding $A_{all} \in \mathbb{R}^{243}$, and passed through a fusion encoder before finally being fed into a classification layer. Motivated by \cite{szegedy2015inception}, each of $\{A_{i1}, A_{i2}, A_{i3}, A_{i4}\}$ is additionally fed into its respective classification layer to accelerate the optimization process. 

Combining all the losses computed from the classification layers, the overall optimization objective becomes 
\begin{equation}
    L = \underset{\theta}{\argmin}\sum_{j=1}^{4}L_{ij} + L_{all},
\end{equation}
where $L_{ij}$ corresponds to the losses computed from $\{A_{i1}, A_{i2}, A_{i3}, A_{i4}\}$, $L_{all}$ corresponds to the loss computed from $A_{all}$, and $\theta$ is the set of learnable parameters.

\vspace{-.2cm}
\paragraph{Training Configuration}
To ensure consistency between Concat-MLP and GaLeNet, we employ Focal Loss \cite{lin2017focalloss} due to its superior performance on imbalanced datasets. We use Adam optimizer \cite{kingma2017adam}, initialized with a learning rate of $1e{-}4$. To combat overfitting, we use Early Stopping \cite{caruana2000earlystopping} with the patience set to 5. Finally, we run each model $5$ times with random initialization of weights and report the averaged metrics.

\section{Results}
\subsection{Single Modalities}
It took some optimization to find the best representation of the image data. Table~\ref{table:single_modality_vis_models_post} shows the performance of the different feature extraction approaches at a fixed scale (Scale-32x). In general, the CLIP representation outperformed both the CNN and the U-Net approaches. This is interesting given that CLIP has not been explicitly trained to detect buildings or damage from satellite images. But, it does have the benefit of being exposed to a much larger variety and number of images during pre-training. We also observe that the performance increases from older to newer versions of CLIP (from top to bottom), which is consistent with the original generalization trend of CLIP models shown by~\cite{openai_2021}. The largest CLIP model (ViT-L/14@336px) performed best and was used for all other experiments that we report.

\begin{table}[ht]
\small
\centering
\caption{Comparison of various feature extractors}
\label{table:single_modality_vis_models_post}
\begin{tabular}{rcccc} 
\toprule
Representation & Bal. Acc. & PR AUC & ROC AUC \\
\midrule
CNN                     & 0.5091 & 0.5711 & 0.8310 \\
U-Net                   & 0.3574 & 0.3970 & 0.6976 \\
CLIP ViT-B/32           & 0.5352 & 0.5925 & 0.8300 \\
CLIP ViT-B/16           & 0.5499 & 0.6023 & 0.8296 \\
CLIP ViT-L/14           & 0.5564 & 0.6107 & 0.8404 \\
CLIP ViT-L/14@336px     & 0.5684 & 0.6183 & 0.8443 \\
\bottomrule
\end{tabular}
\end{table}

Table~\ref{table:single_modality_vis_crops_post} presents the performance metrics of CLIP ViT-L/14@336px embeddings obtained using various cropping strategies on pre- and post-disaster satellite images. Interestingly, we notice a difference between the pre- and post-disaster representations. In the pre-disaster case, the performance improves progressively as we ``zoom out'' from the building. However, in the post-disaster case the opposite is true and ``zooming in'' to the building seems to boost performance. We reason that this is because in the post-disaster case the direct visual evidence of the building damage is important to assessing damage severity. However, in the pre-disaster case this is not available, so the model relies on inferring potential damage using the contextual information surrounding the building. This might include whether it is sheltered by other buildings or forests, or whether it is close to loose structures that could get uplifted and blown into the building.

In both cases, the highest performance is achieved by concatenating all the different scales, implying that both pre- and post-disaster damage assessment benefit from combining information obtained at multiple scales.

Table~\ref{table:multimodal-galenet} (rows 1-4) show the performance of each modality in the unimodal setting using LogReg as the model. The ROC AUC performance shows that on their own, each modality provides some utility for assessing building damage severity, even for proactive modalities that do have access to direct visual evidence of the damage. As expected, the post-disaster image modality provides a significant boost in performance over the proactive modalities.

\begin{table*}[ht]
\small
\centering
\caption{Comparison of various cropping strategies and performance for pre- and post-disaster images.}
\label{table:single_modality_vis_crops_post}
\begin{tabular}{cccccccc} 
\toprule
\multicolumn{1}{c}{} & \multicolumn{3}{c}{Pre-disaster}& \multicolumn{3}{c}{Post-disaster} \\
Image Scale & Bal. Acc. & PR AUC & ROC AUC & Bal. Acc. & PR AUC & ROC AUC \\
\midrule
Scale-1x   & \textbf{0.4963} & \textbf{0.5329} & \textbf{0.7878} & 0.4929 & 0.5598 & 0.8129 \\
Scale-4x   & 0.4835 & 0.5191 & 0.7797 & 0.5147 & 0.5804 & 0.8229 \\
Scale-16x  & 0.4762 & 0.5201 & 0.7799 & 0.5381 & 0.6083 & 0.8386 \\
Scale-32x  & 0.4588 & 0.4944 & 0.7620 & \textbf{0.5684} & \textbf{0.6183} & \textbf{0.8443} \\
\midrule
All Scales  & \textbf{0.4931} & \textbf{0.5439} & \textbf{0.7962} & \textbf{0.5707} & \textbf{0.6430} & \textbf{0.8570} \\
\bottomrule
\end{tabular}
\end{table*}

\begin{table*}[ht]
\small
\centering
\caption{Comparison of unimodal and multimodal baselines with GaLeNet, for the modalities: weather data (W), hurricane trajectory (T), pre-disaster image (Pre), and post-disaster image (Post).}
\label{table:multimodal-galenet}
\begin{tabular}{rrcccc} 
\toprule
Model & Features & Scenario & Bal. Acc. & PR AUC & ROC AUC \\
\midrule
LogReg              & W            & Proactive & 0.5238 & 0.4369 & 0.7122  \\
LogReg              & T            & Proactive & 0.4348 & 0.5145 & 0.7508  \\
LogReg              & Pre          & Proactive & 0.4931 & 0.5439 & 0.7962  \\
LogReg              & Post         & Reactive  & 0.5707 & 0.6430 & 0.8570  \\
\midrule
LogReg              & W + T + Pre  & Proactive & 0.5110 & 0.5533 & 0.8090  \\
Concat-MLP          & W + T + Pre  & Proactive & 0.6357 & 0.5518 & 0.8072  \\
GaLeNet             & W + T + Pre  & Proactive & \textbf{0.6495} & \textbf{0.5645} & \textbf{0.8140}  \\
\midrule
LogReg              & W + T + Post & Reactive & 0.5773 & 0.6472 & 0.8631  \\
Concat-MLP          & W + T + Post & Reactive & 0.6798 & 0.6556 & 0.8648  \\
GaLeNet             & W + T + Post & Reactive & \textbf{0.6875} & \textbf{0.6680} & \textbf{0.8732} \\
\bottomrule
\label{table:res-multimodal}
\end{tabular}
\end{table*}

\begin{figure*}[ht]
    \centering
    \begin{subfigure}{.32\textwidth}
        \centering
        \includegraphics[width=\linewidth]{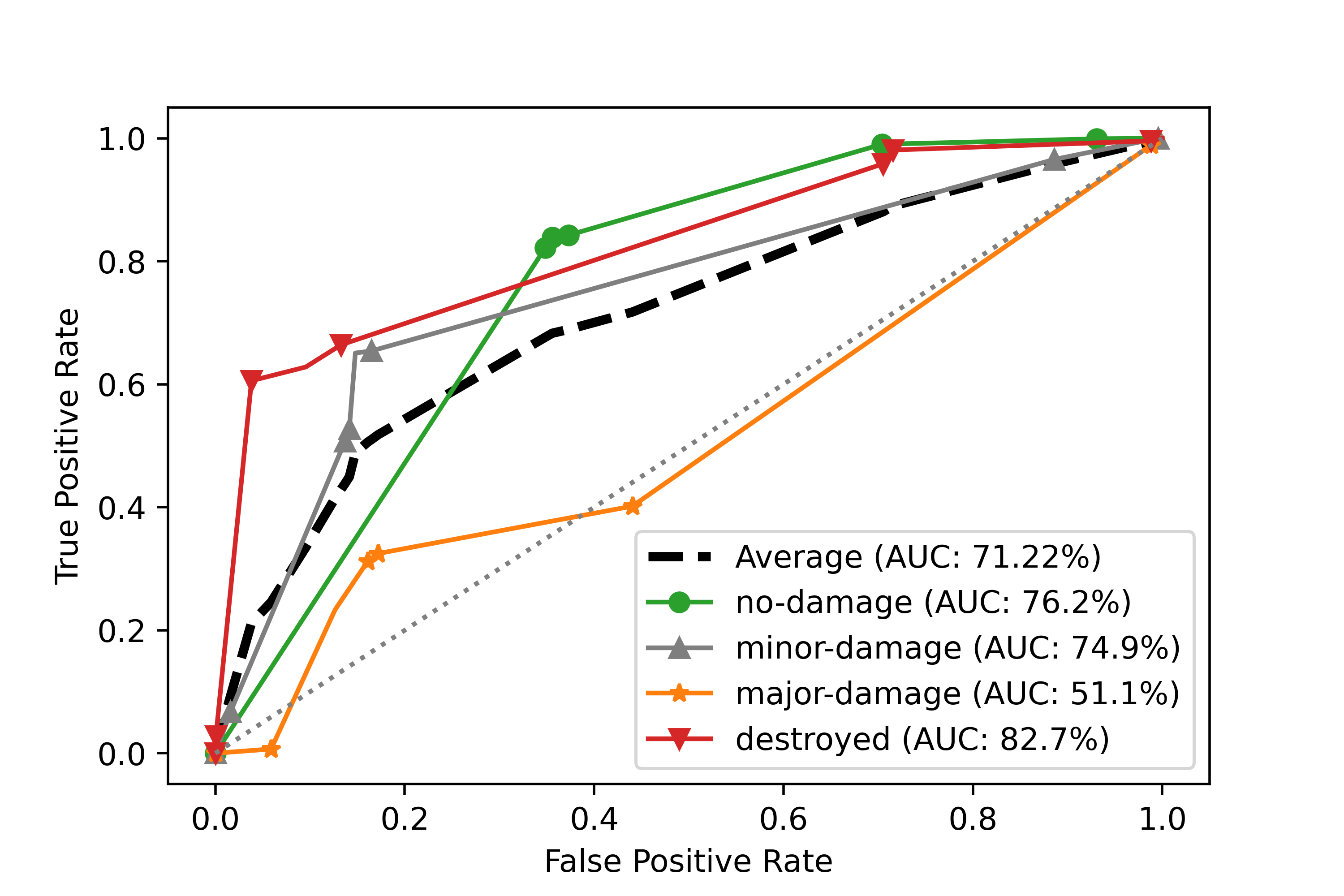}
        \caption{Weather (W)}
    \end{subfigure}
    \begin{subfigure}{.32\textwidth}
        \centering
        \includegraphics[width=\linewidth]{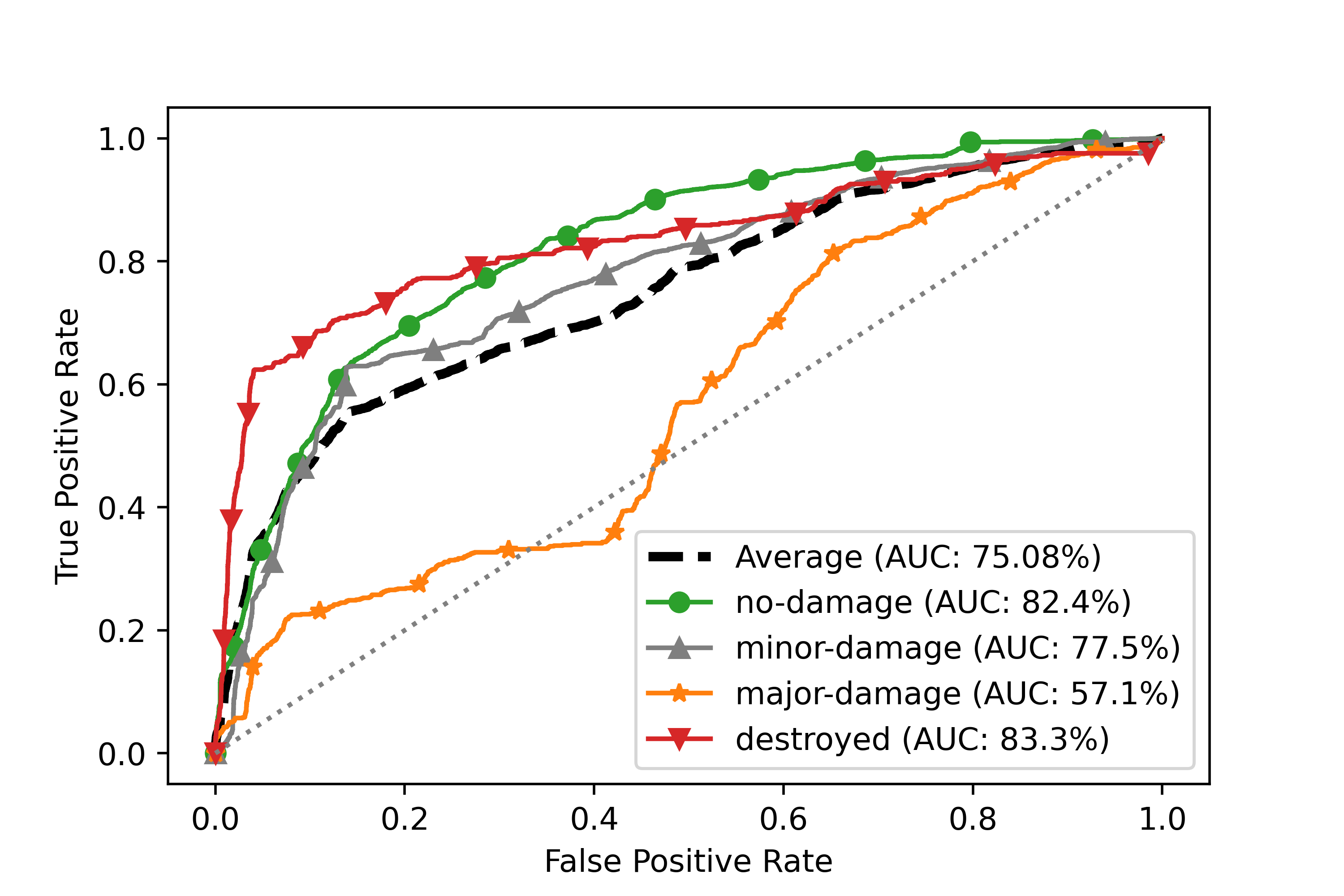}
        \caption{Hurricane Trajectory (T)}
    \end{subfigure}
    \begin{subfigure}{.32\textwidth}
        \centering
        \includegraphics[width=\linewidth]{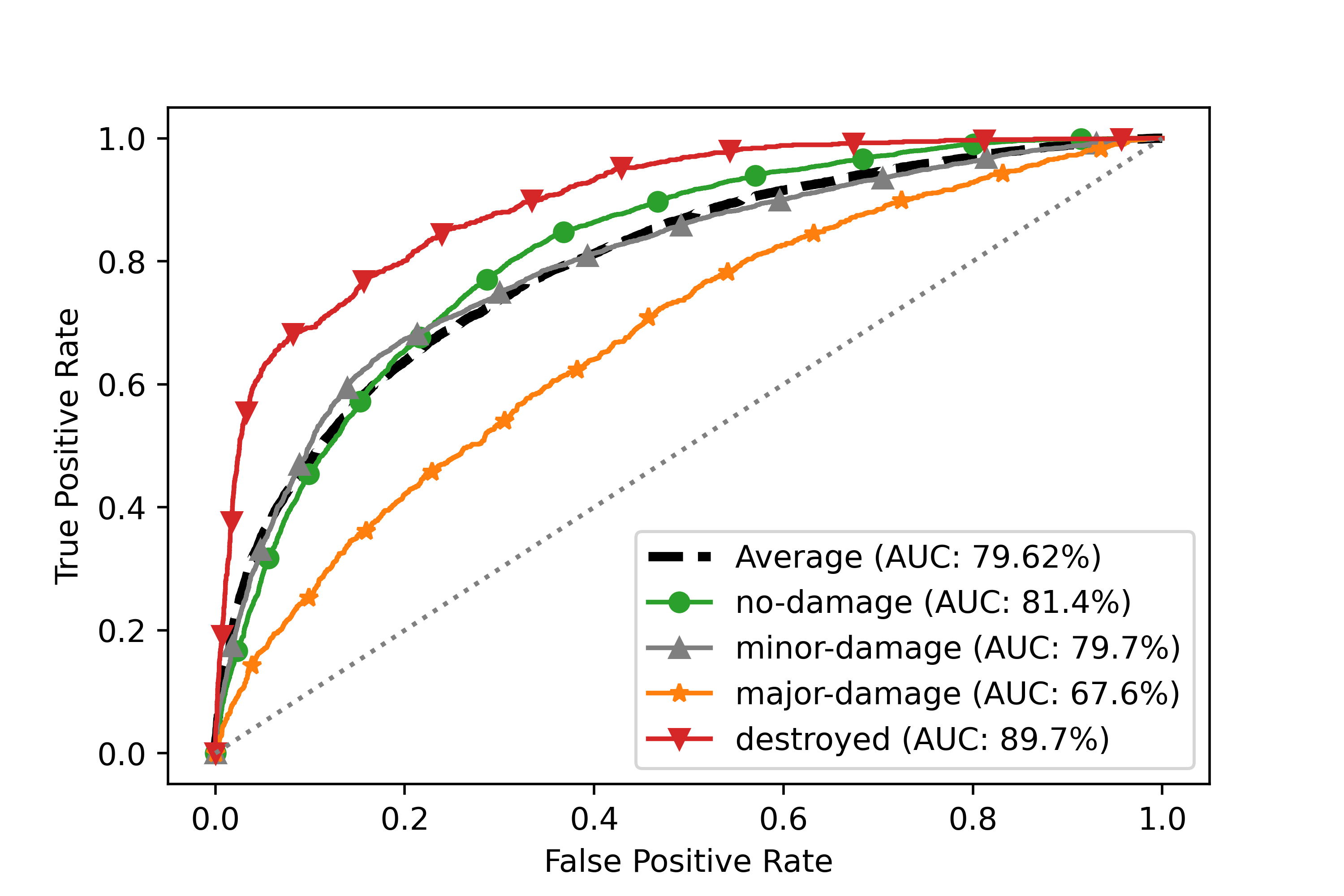}
        \caption{Pre-disaster Images (Pre)}
    \end{subfigure}
    \begin{subfigure}{.32\textwidth}
        \centering
        \includegraphics[width=\linewidth]{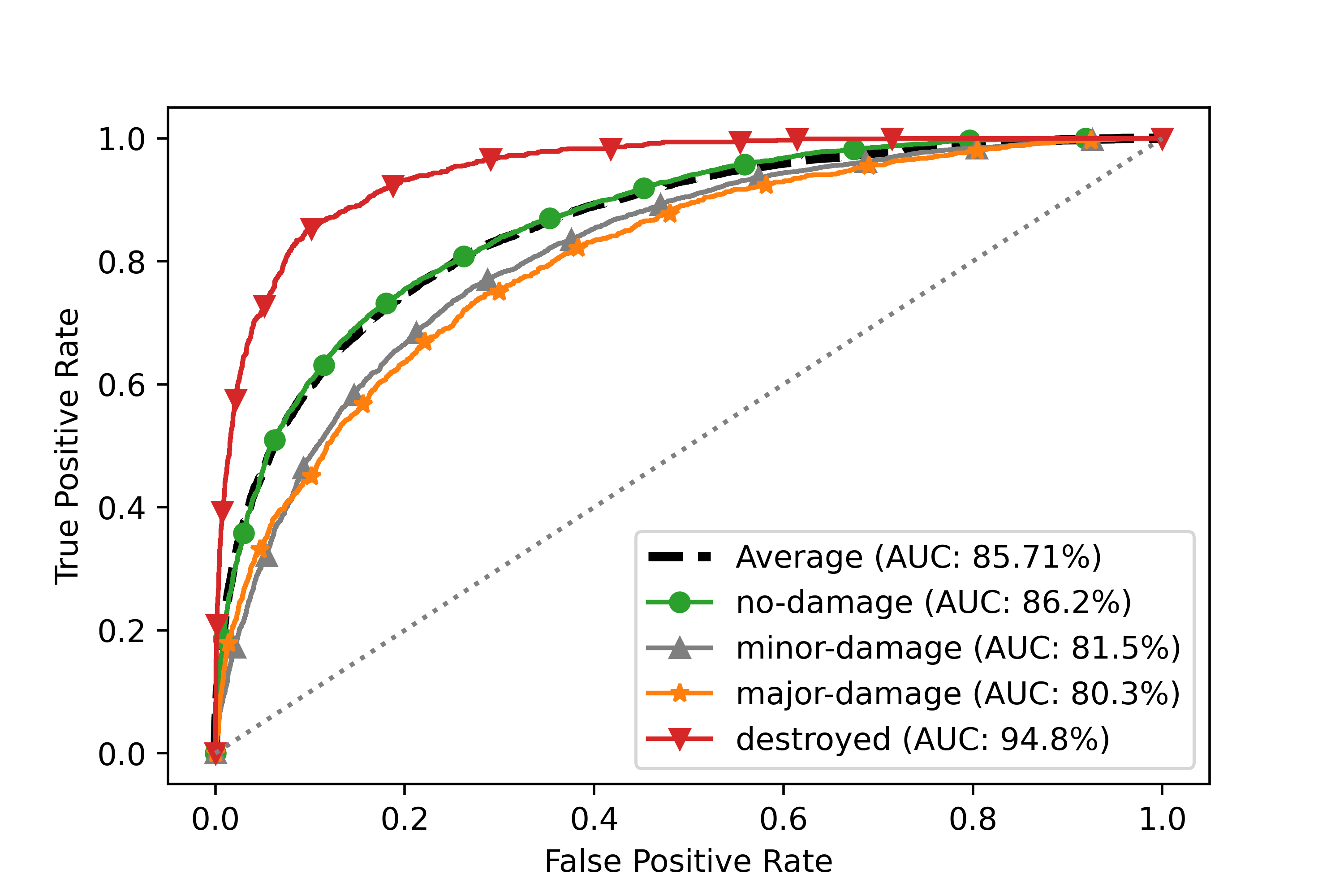}
        \caption{Post-disaster Images (Post)}
    \end{subfigure}
    % \rulesep
    \begin{subfigure}{.32\textwidth}
        \centering
        \includegraphics[width=\linewidth]{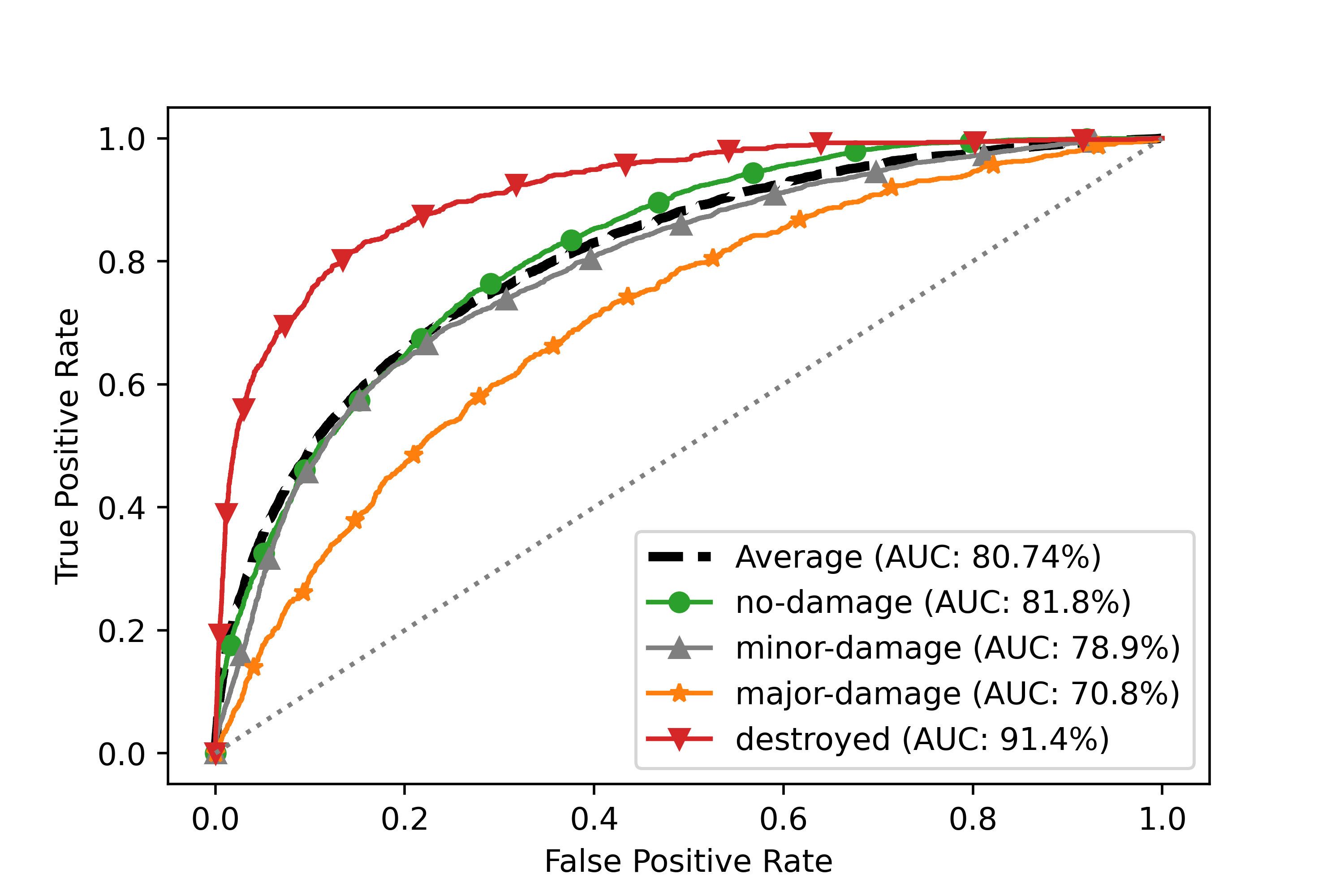}
        \caption{GaLeNet Proactive}
    \end{subfigure}
    \begin{subfigure}{.32\textwidth}
        \centering
        \includegraphics[width=\linewidth]{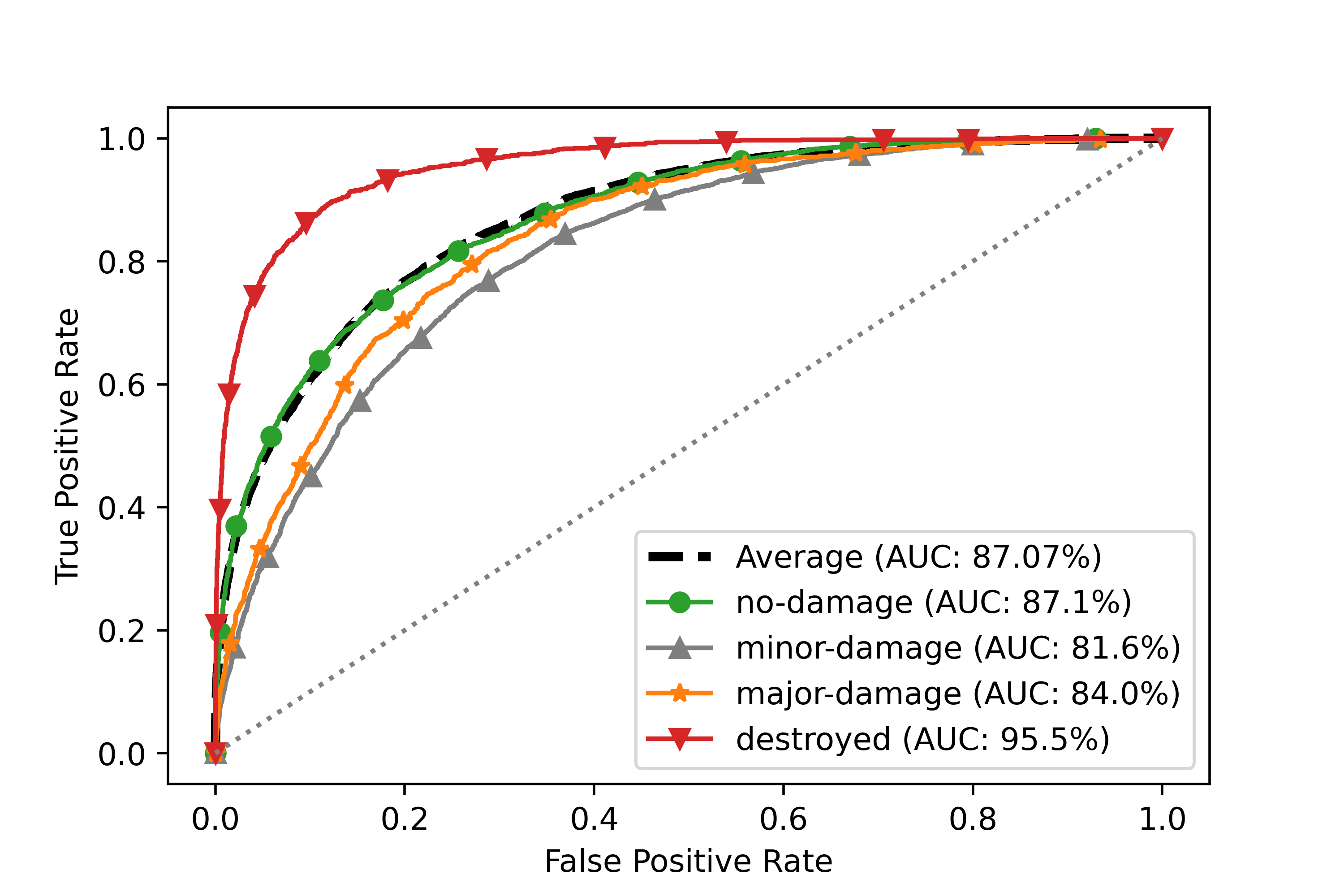}
        \caption{GaLeNet Reactive}
    \end{subfigure}
    \caption{Comparison of ROC curves between unimodal baselines (a-d) and GaLeNet used for the proactive (e) and reactive (f) use cases.}
\label{figure:roc-auc}
\end{figure*}

Figure~\ref{figure:roc-auc}(a-d) shows ROC curves for the unimodal baselines. It is evident that for the proactive modalities (a-c), there is a noticeable drop in performance for the ``Major Damage'' class, however this is not the case for the post-disaster image modality. When studying the class predictions of the model, it is evident that the proactive models tend to confuse ``Major Damage'' with ``Destroyed''. In the post-disaster case, the model becomes much better at distinguishing the different levels of damage severity as it has direct visual evidence of the damage.

\subsection{Multiple Modalities}
\label{sec:results-multimodal}
Table~\ref{table:multimodal-galenet} (rows 5-10) provides a comparison of our proposed framework (GaLeNet) to the LogReg and Concat-MLP baselines for both the proactive and reactive scenarios. 

Firstly, we observe that the LogReg is able to benefit from using all input modalities, as the multimodal LogReg outperforms all of the unimodal LogReg models in both the proactive and reactive cases. This implies that the different modalities have some complimentary information for inferring the damage severity, even if provided with post-disaster images.

Secondly, we note that a naïve neural network approach (Concat-MLP) does not offer much benefit over the LogReg, as it does not consistently outperform the LogReg in either the proactive or reactive cases.

GaLeNet outperforms both fusion baselines across all metrics for both scenarios. Focusing on the Proactive use-case, GaLeNet achieves an increase of $\thicksim$14\% in Bal. Acc. over LogReg. In the reactive use case, GaLeNet achieves similar performance boost in Bal. Acc.

\begin{figure*}[htbp]
    \centering
    \begin{subfigure}{.4\textwidth}
        \centering
        \includegraphics[width=\linewidth]{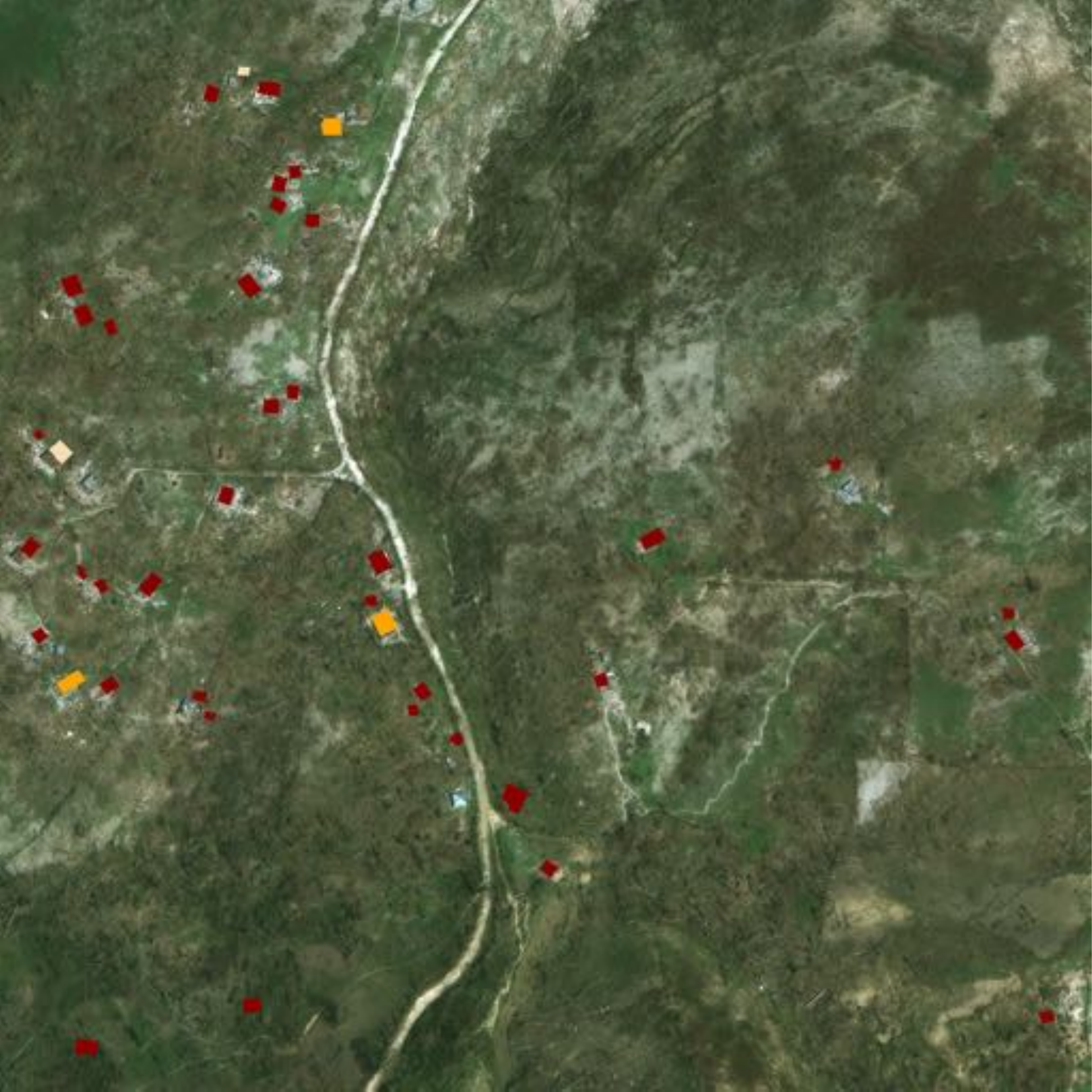}
        \caption{Ground truth - Hurricane Matthew}
    \end{subfigure}
    \begin{subfigure}{.4\textwidth}
        \centering
        \includegraphics[width=\linewidth]{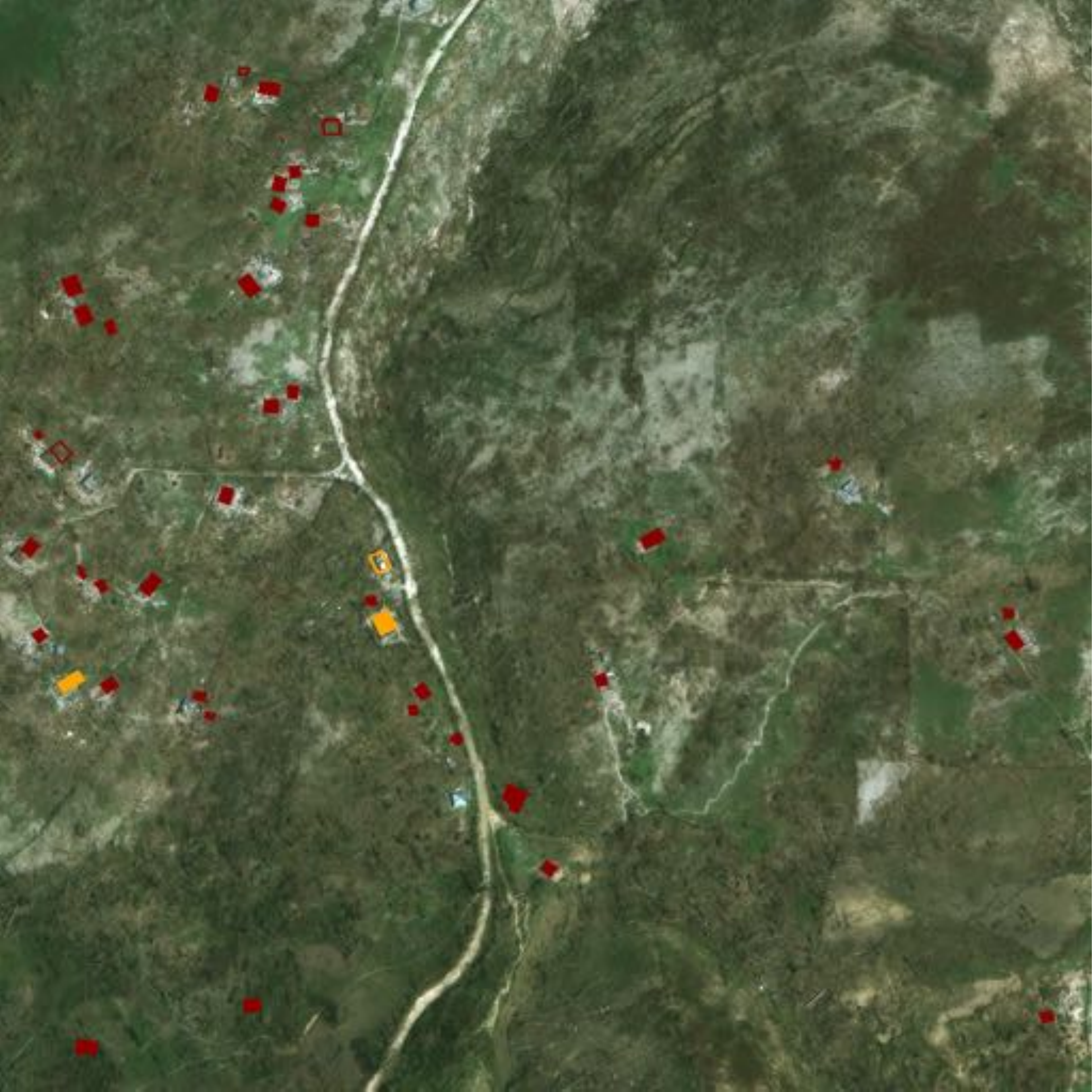}
        \caption{GaLeNet proactive - Hurricane Matthew}
    \end{subfigure}\\
    \begin{subfigure}{.4\textwidth}
        \centering
        \includegraphics[width=\linewidth]{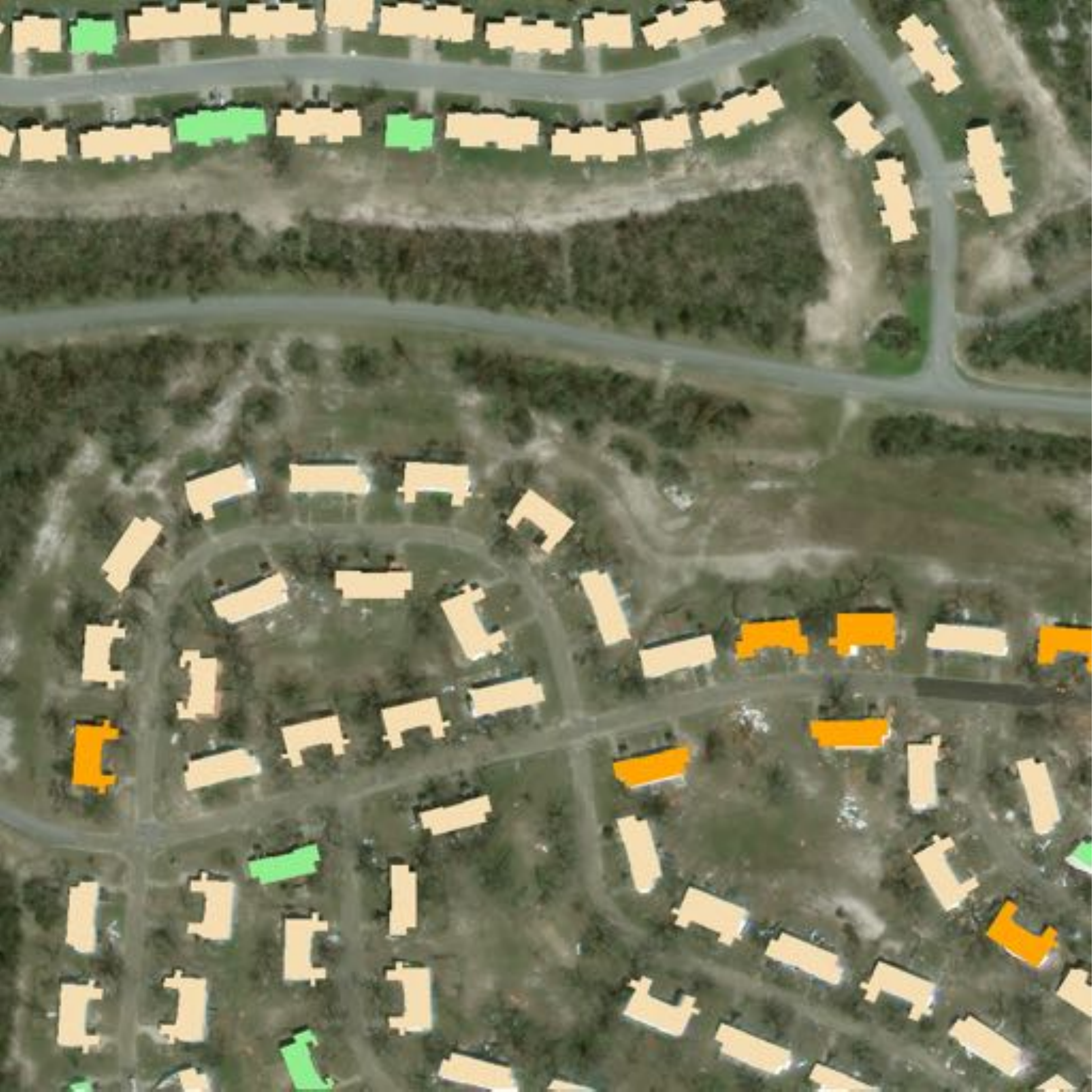}
        \caption{Ground truth - Hurricane Michael}
    \end{subfigure}
    \begin{subfigure}{.4\textwidth}
        \centering
        \includegraphics[width=\linewidth]{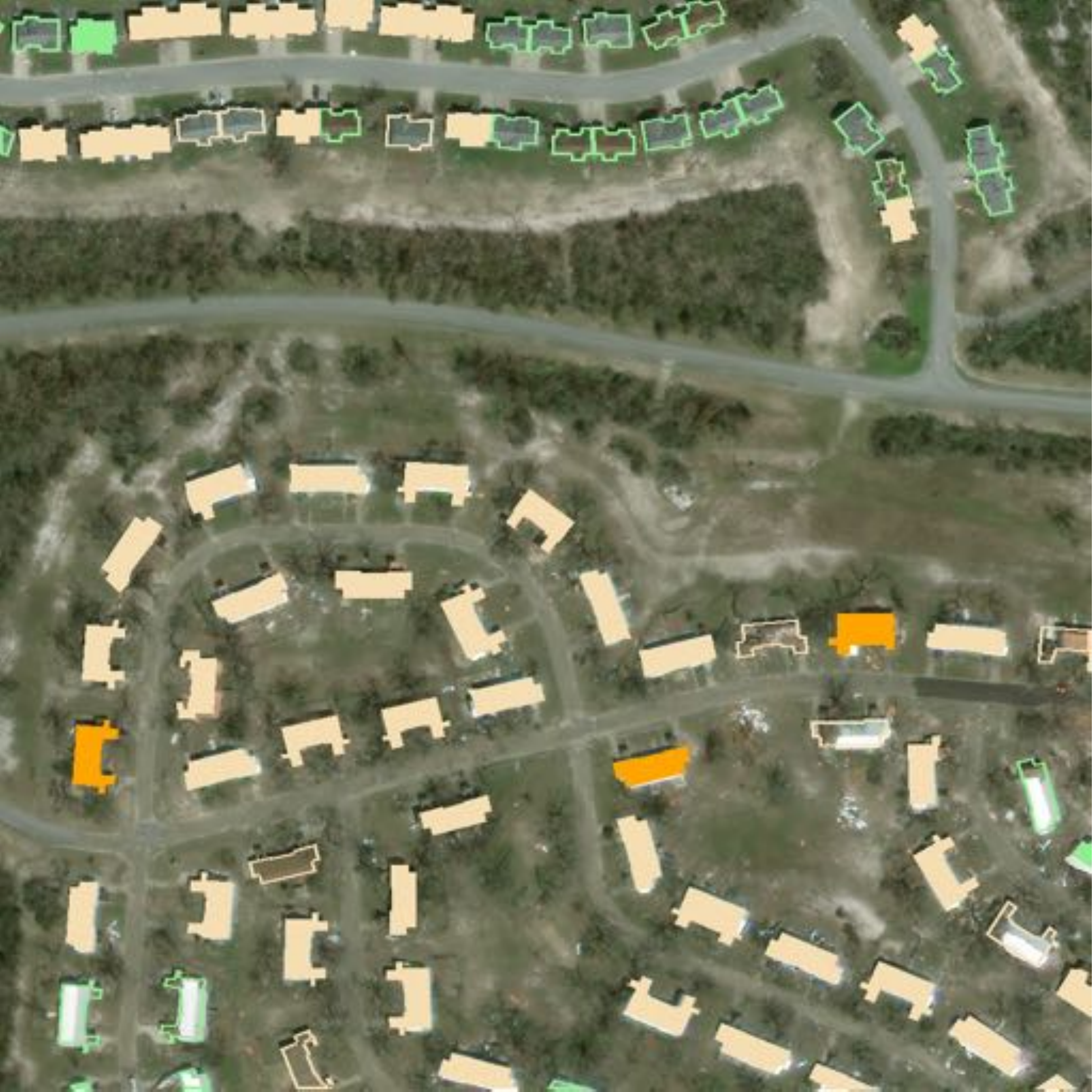}
        \caption{GaLeNet proactive - Hurricane Michael}
    \end{subfigure}
    \caption{Visual comparison between ground truth labels and GaLeNet predictions for the \emph{proactive} scenario. Images from hurricanes Matthew and Michael are shown in (a, b) and (c, d), respectively. ``No Damage``, ``Minor Damage'', ``Major Damage'' and ``Destroyed'' are shown in green, wheat, orange and red, respectively. Correct predictions are filled in the relevant colour, whereas misclassified predictions are outlined in the colour of the predicted label. Note that the building masks are taken from the xBD dataset and not predicted by the model, they are used only for visualization purposes.}
\label{figure:vis-proactive}
\end{figure*}

\begin{figure*}[htbp]
    \centering
    \begin{subfigure}{.4\textwidth}
        \centering
        \includegraphics[width=\linewidth]{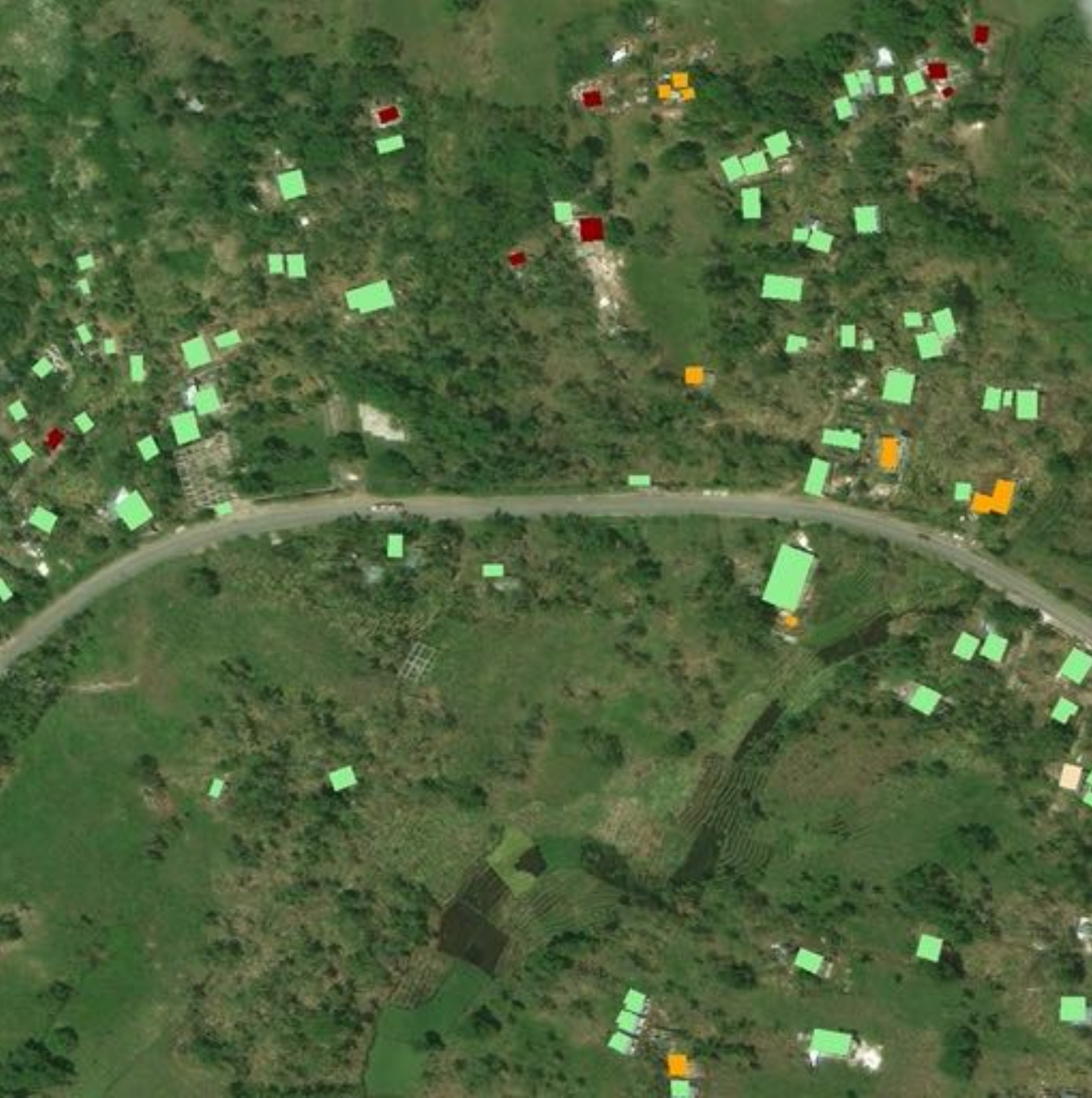}
        \caption{Ground truth - Hurricane Matthew}
    \end{subfigure}
    \begin{subfigure}{.4\textwidth}
        \centering
        \includegraphics[width=\linewidth]{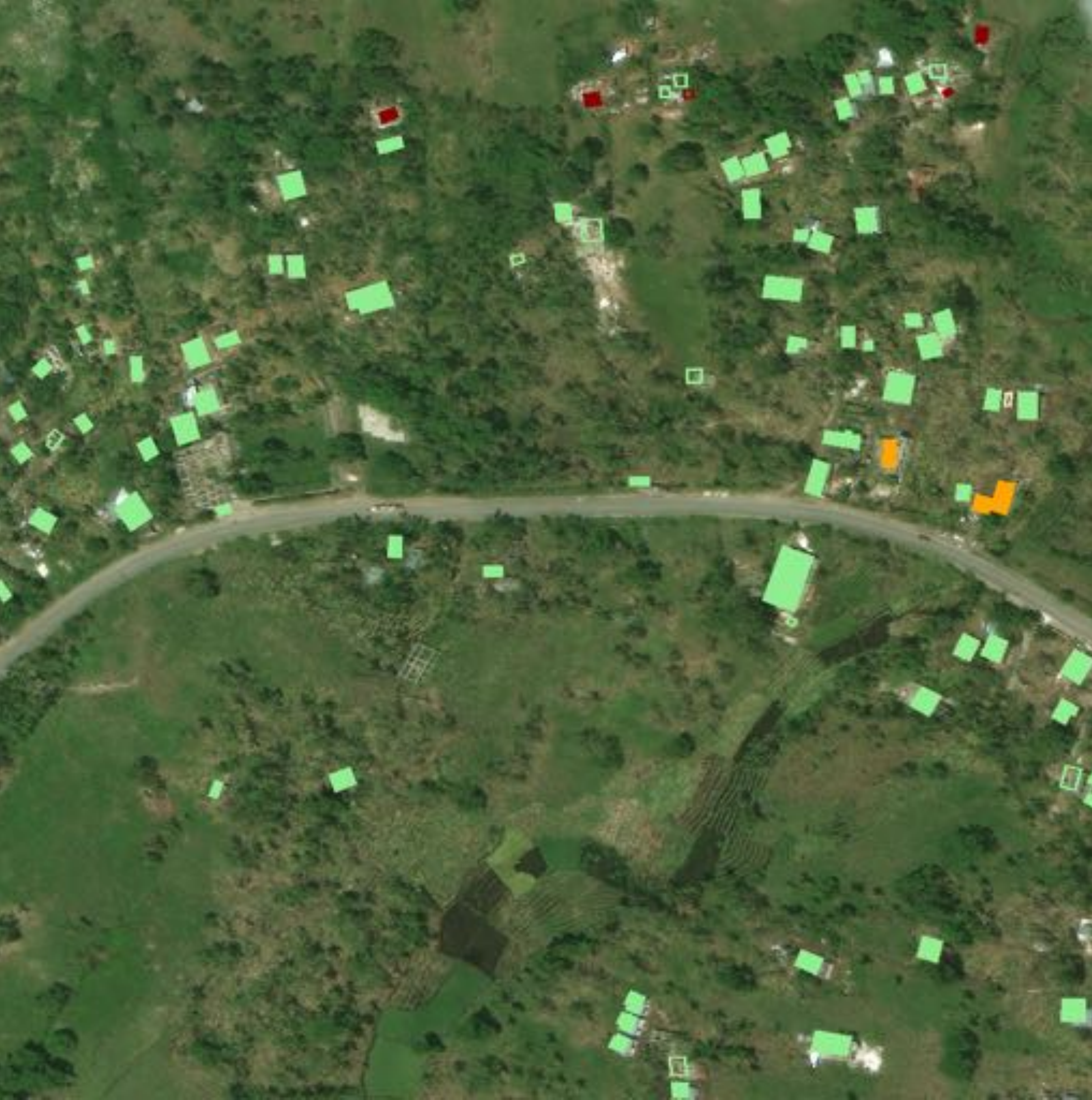}
        \caption{GaLeNet reactive - Hurricane Matthew}
    \end{subfigure}
    \begin{subfigure}{.4\textwidth}
        \centering
        \includegraphics[width=\linewidth]{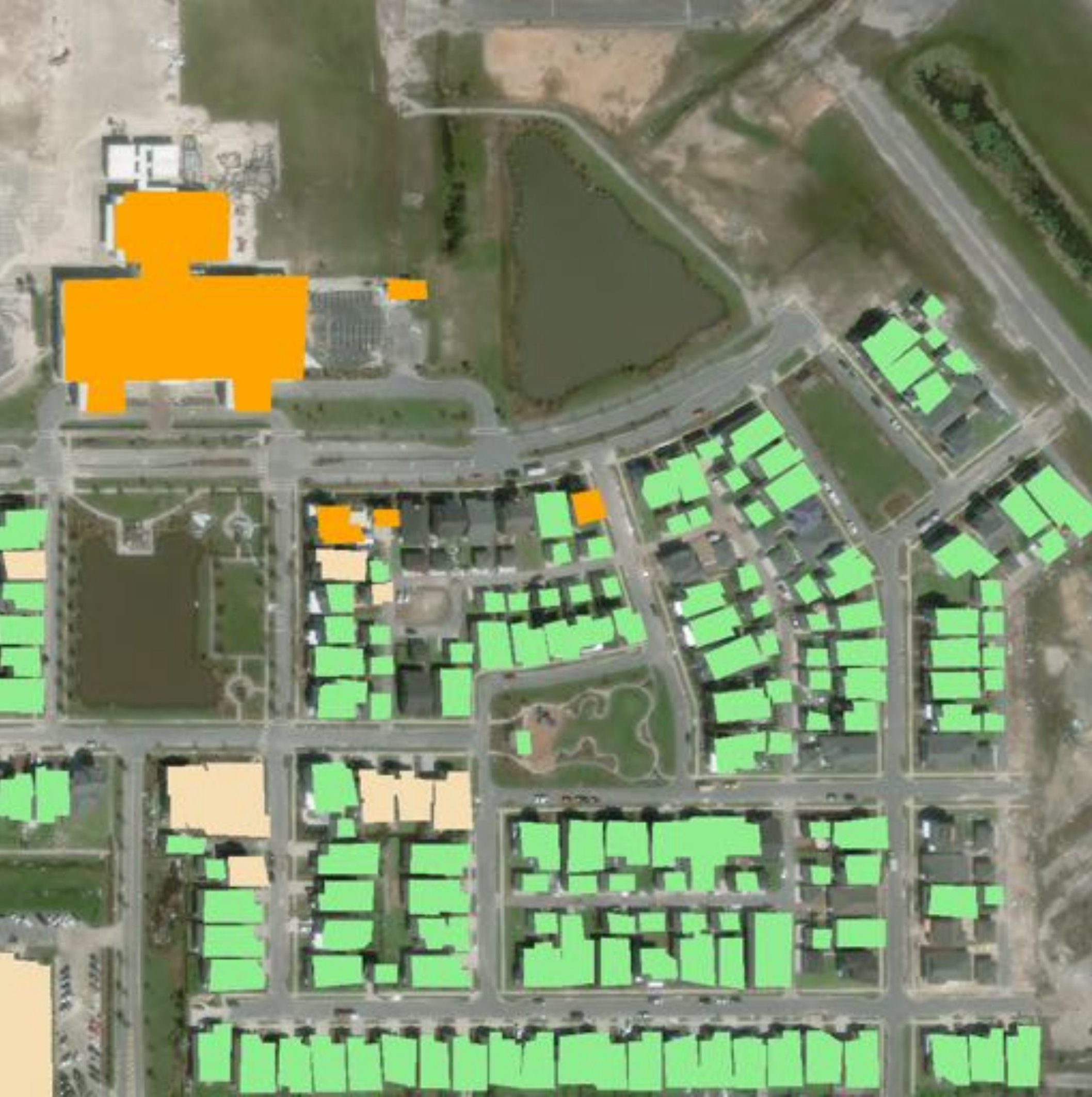}
        \caption{Ground truth - Hurricane Michael}
    \end{subfigure}
    \begin{subfigure}{.4\textwidth}
        \centering
        \includegraphics[width=\linewidth]{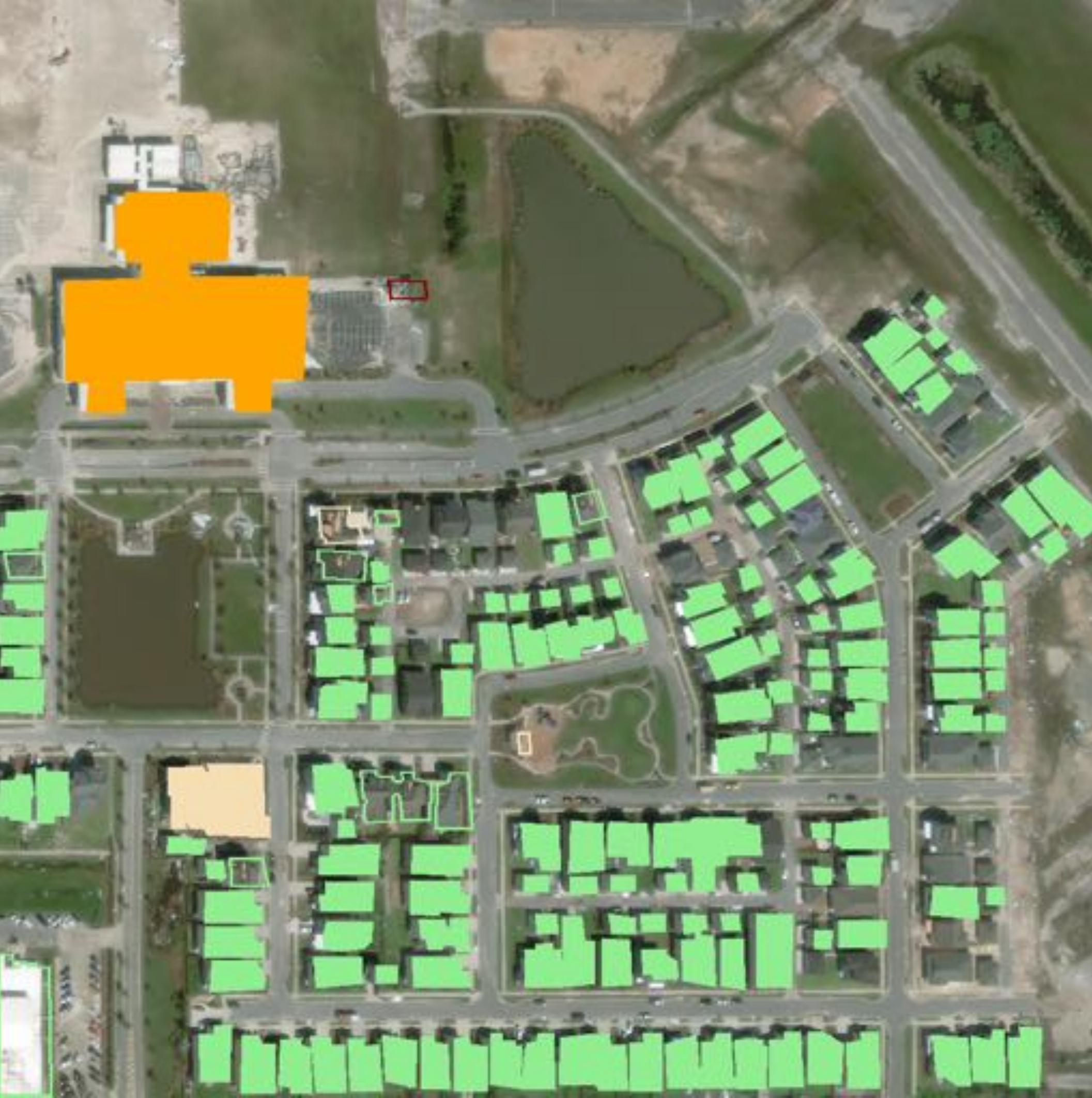}
        \caption{GaLeNet reactive - Hurricane Michael}
    \end{subfigure}
    \caption{Visual comparison between ground truth labels and GaLeNet predictions for the \emph{reactive} scenario. Images from hurricanes Matthew and Michael are shown in (a, b) and (c, d), respectively. ``No Damage``, ``Minor Damage'', ``Major Damage'' and ``Destroyed'' are shown in green, wheat, orange and red, respectively. Correct predictions are filled in the relevant colour, whereas misclassified predictions are outlined in the colour of the predicted label. Note that the building masks are taken from the xBD dataset and not predicted by the model, they are used only for visualization purposes.}
\label{figure:vis-reactive}
\end{figure*}

Qualitatively, GaLeNet demonstrates its capability to correctly identify the severity of damage across buildings of multiple sizes (Figures~\ref{figure:vis-proactive} and \ref{figure:vis-reactive}). This can be attributed to the availability of visual features extracted at different image scales and the normalization of the image scale. 

Finally, it is worth noting that GaLeNet outperforms its closest architectural neighbor, Concat-MLP, across all metrics while having only 189K parameters ($52.75\%$ fewer parameters compared to Concat-MLP). This further attests to the effectiveness of GaLeNet's compact architectural design. 

\section{Limitations \& Future Work}
During the process of this work, we found several limitations which open up impactful avenues of future work:
\begin{itemize}
    \item We experimented with additional modalities such as elevation data and twitter data, and while these gave above chance performance in an unimodal setting, they did not offer any benefit in a multimodal setting. This could be due to the limited size of our dataset, we had to use a relatively shallow model which was unable to reason across these additional modalities. In addition, the twitter dataset was relatively sparse, and the resultant model suffered from ``modality dropout''. We feel that a solution to this is essential to real-world application of such a model.
    \item Several attempts were made to improve the representation of the weather data and the hurricane trajectory data, however, it was found that only relatively simple featurizations of each offered the best performance. We attribute this to the granularity (in terms of longitude and latitude) of the data and to the overall size of the dataset and the inability of the model to learn a more nuanced representation of the data. The dataset size also prevented us experimenting with different co-learning strategies.
    \item Following error analysis, we realized that our framework struggles to identify damage on buildings that are circular. This can be attributed to their rare occurrence in the real world, and consequently in the dataset. Future work can, therefore, explore extending existing datasets to cover edge cases.
    \item The dataset only contained two hurricanes, so we were unable to test generalization to new natural disasters. Collecting data and testing across more natural disasters is an essential next step (in our roadmap) before these models can be utilized in the real world.
\end{itemize}

% =================================================
% Conclusion
% =================================================
\section{Conclusion}
We proposed a multimodal framework, GaLeNet, that is capable of predicting the severity of damage to buildings after a hurricane even if it does not have access to direct visual evidence of the damage. Moreover, when such information is available, GaLeNet is able to use it to increase the accuracy of its predictions. Through extensive evaluation with data from two hurricanes, we show the effectiveness of GaLeNet by comparing it against multiple unimodal and multimodal baselines. 

We believe that such a framework could provide crucial early insights and intelligence on the natural disaster before the true damage is known. Thus, it could help allocate resources for disaster relief where they matter most, and update this allocation as more data becomes available.

In addition, it is possible that proactive data such as weather predictions, predicted hurricane trajectories and ``blue skies'' imaging could provide useful insights about the risk of particular buildings to damage in the event of a natural disaster. Such actionable insights could be used to strengthen our defenses against disasters before they occur.

As climate change and its effects become more pronounced, the need to address its challenges has never been more pressing. We hope that GaLeNet, despite its limitations, can ignite future research and development in this direction.

\clearpage
\clearpage 
{\small
\bibliographystyle{ieee_fullname}
\bibliography{egbib}
}

\end{document}